\pdfoutput=1

\documentclass[11pt]{article}
\usepackage{color}
\usepackage{times}
\usepackage{epsfig}
\usepackage{graphicx}
\usepackage{amsmath}
\usepackage{amsthm}
\usepackage{amssymb}
\usepackage{mathtools}
\usepackage{multirow}
\usepackage{bbm}
\usepackage{dsfont}

\usepackage[table, dvipsnames]{xcolor}

\usepackage[utf8]{inputenc}
\usepackage{booktabs,siunitx}
\usepackage{multirow}
\usepackage{multicol}
\usepackage{amsmath}
\usepackage{subcaption}
\usepackage{caption}
\usepackage{graphicx}
\usepackage{pifont}

\usepackage{tikz}
\usetikzlibrary{shapes,arrows,patterns}
\usetikzlibrary{positioning}
\usetikzlibrary{calc}

\newtheorem{theorem}{Theorem}[section]

\newcommand{\cut}[1]{}

\definecolor{red}{RGB}{255, 117, 115}
\definecolor{green}{RGB}{171, 255, 175}

\usepackage{ACL2023}

\usepackage{times}
\usepackage{latexsym}

\usepackage[T1]{fontenc}

\usepackage[utf8]{inputenc}

\usepackage{microtype}

\usepackage{inconsolata}

\definecolor{cadmiumgreen}{rgb}{0.0, 0.42, 0.24}
\definecolor{cardinal}{rgb}{0.77, 0.12, 0.23}
\definecolor{cadmiumred}{rgb}{0.89, 0.0, 0.13}
\newcommand*{\promptfont}{\fontfamily{courier}\selectfont}
\newtheorem{conjecture}[theorem]{Conjecture}

\title{
Large Language Models Sensitivity to The Order of Options in Multiple-Choice Questions}

\author{Pouya Pezeshkpour \\
  Megagon Labs \\
  \texttt{pouya@megagon.ai} \\\And
  Estevam Hruschka\\
    Megagon Labs \\
  \texttt{estevam@megagon.ai}
}

\begin{document}
\maketitle
\begin{abstract}
Large Language Models (LLMs) have demonstrated remarkable capabilities in various NLP tasks. 
However, previous works have shown these models are sensitive towards prompt wording, and few-shot demonstrations and their order, posing challenges to fair assessment of these models. 
As these models become more powerful, it becomes imperative to understand and address these limitations.  
In this paper, we focus on LLMs robustness on the task of multiple-choice questions---commonly adopted task to study reasoning and fact-retrieving capability of LLMs. 
Investigating the sensitivity of LLMs towards the order of options in multiple-choice questions,  
we demonstrate a considerable performance gap of approximately $13\%$ to $75\%$ in LLMs on different benchmarks, when answer options are reordered, even when using demonstrations in a few-shot setting. 
Through a detailed analysis, we conjecture that this sensitivity arises when LLMs are uncertain about the prediction between the top-2/3 choices, and specific options placements may favor certain prediction between those top choices depending on the question caused by positional bias. 
We also identify patterns in top-2 choices that amplify or mitigate
the model's bias toward option placement. 
We found that for amplifying bias, the optimal strategy involves positioning the top two choices as the first and last options. Conversely, to mitigate bias, we recommend placing these choices among the adjacent options. 
To validate our conjecture, we conduct various experiments and adopt two approaches to calibrate LLMs' predictions, leading to up to $8$ percentage points improvement across different models and benchmarks. 
\end{abstract}

\section{Introduction}
Large Language Models (LLMs) have demonstrated impressive performance on various tasks, surpassing that of supervised models and, in some cases, even outperforming humans \citep{chowdhery2022palm,touvron2023llama,openai2023gpt-4}. 
However, despite their impressive capabilities, previous research has highlighted certain limitations. For instance, LLMs have shown significant sensitivity to small changes in the prompt \citep{zhao2021calibrate,wang2023adversarial,zhu2023promptbench}.  
Therefore, a more comprehensive and conclusive analysis of different aspects that can affect/limit LLMs' performance is crucial for a fair assessment and their successful real-world adoption.

One significant limitation lies in the robustness of LLMs concerning the arrangement of various components in a prompt, as it directly impacts the assessment of their capability in understanding and reasoning for specific tasks. 
Prior research has demonstrated that LLMs exhibit sensitivity to the arrangement of few-shot demonstrations \citep{zhao2021calibrate} and the order of appearance for responses generated by candidate models when LLMs are used as referees to evaluate quality \citep{wang2023large}.
Given these findings, it becomes pertinent to inquire whether LLMs are also sensitive to the order of elements of the prompts in different tasks. For example, how much does the order of options in multiple-choice question (MCQ) answering tasks can impact the LLMs performance.

\begin{figure}[t]
    \centering
    \includegraphics[width=0.98\columnwidth]{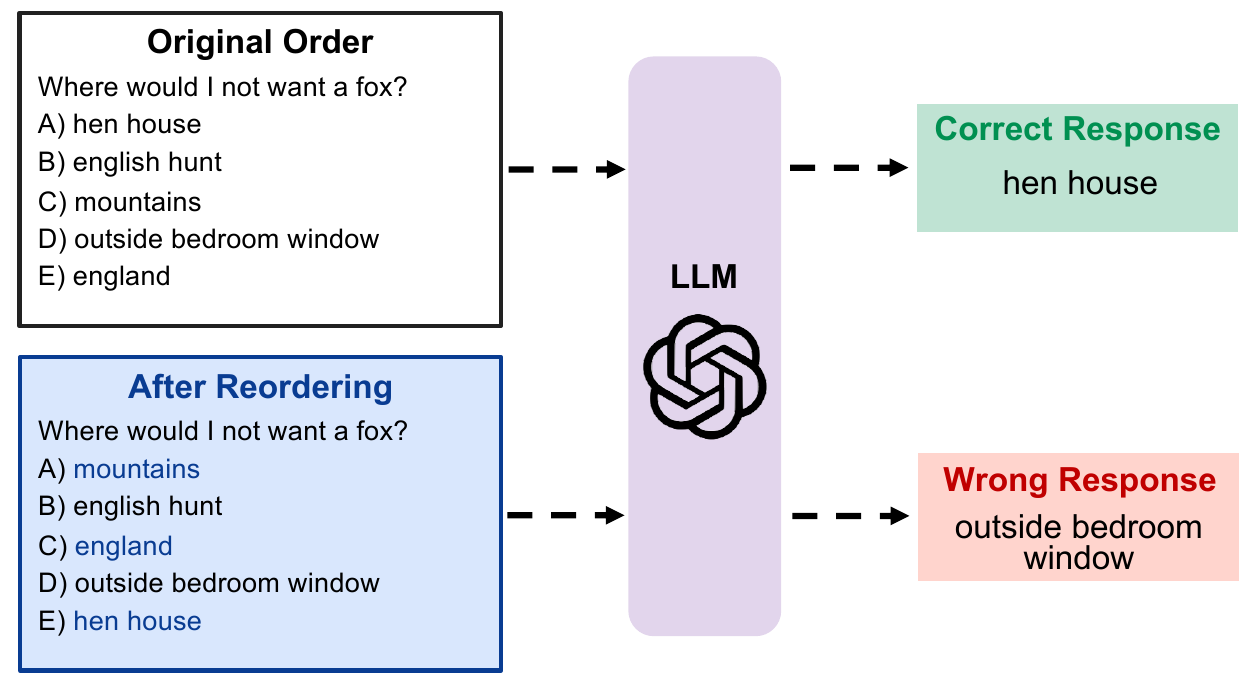}
    \caption{\textbf{GPT-4 sensitivity to reordering options:} upon changing the order of choices, GPT-4 change its prediction from ``hen house'' to ``outside of bedroom window'' (the example is from CSQA dataset).}
    \label{fig:over}
\end{figure}

In this paper, we investigating the sensitivity of LLMs to the order of options in multiple-choice questions; using it as a proxy to understand LLMs sensitivity to the order of prompt elements in instruction- or demonstration-based paradigm. 
We demonstrate an example of GPT-4's sensitivity to options order in Figure \ref{fig:over}, using a sample from the CSQA benchmark \citep{talmor2018commonsenseqa}. 
Notably, by merely rearranging the placement of options among choices A, C, and E, GPT-4 incorrectly predicts the answer to be ``outside bedroom window''. 
Within this context, we aim to address the following research questions: 
(1) To what extent do LLMs exhibit sensitivity to the order of options in multiple-choice questions? 
(2) What factors contribute to LLMs' sensitivity to the order of options? 
(3) How can we improve LLMs' robustness to the order of options?

To answer the first question, we conducted experiments using GPT-4 \citep{openai2023gpt-4} and InstructGPT (text-davinci-003) \citep{ouyang2022training} on five different multiple-choice question benchmarks. 
Surprisingly, we discovered a substantial sensitivity gap of up to $75\%$ in the zero-shot setting. 
Additionally, in the few-shot setting, we observed that introducing demonstrations to the prompt only led to marginal improvements in LLMs' robustness if their performance increased. 
Moving on to the second question, we put forth a conjecture that the sensitivity of LLMs stems from their positional bias, wherein they tend to favor certain placements when uncertain about the answer among the top choices. 
To validate our conjecture, we analyzed instances where the models' predictions changed upon reordering the options. 
Furthermore, we demonstrated that the complexity of the number of choices, while retaining the top possible answers, had only a gradual impact (if any) on performance improvement.

Additionally, we discerned patterns in the occurrence of top-2 possible choices that influence the model's probability of selecting a particular option or somewhat mitigate LLMs' positional bias. For amplifying bias, we found that the optimal strategy involves positioning the top two choices as the first and last options. Conversely, to mitigate bias, we recommend placing these choices among the adjacent options. 
To validate our findings, we conducted qualitative evaluations. 
Addressing the last question, we demonstrated that employing two different calibrating approaches led to a notable improvement in LLMs' performance, up to $8$ percentage points.  
Through these investigations, we contribute to a deeper understanding of how the order of options affects LLMs' decision-making in multiple-choice questions (MCQ) and offer practical solutions, which go beyond simple bootstrapping, to increase their robustness and accuracy in such scenarios.

\section{Background and Experimental Details}
This paper focuses on the task of multiple-choice question answering. In multiple-choice questions, the objective is to identify the correct answer to a given question from a set of possible options (an illustration is presented in Figure \ref{fig:over}). To address this task using in-context learning models, we present a prompt in the following format: {\promptfont ``Choose the answer to the question only from A, B, C, D, and E choices. Question: \{question\}. Choices: \{options\}. Answer:''} to the models. 
Also, for few-shot demonstrations we randomly choose 100 samples and draw demonstrations from those samples.

\paragraph{Models:} 
In our evaluations, we employed two widely-used large language models, InstructGPT (text-davinci-003) \citep{ouyang2022training} and GPT-4 \citep{openai2023gpt-4}, to represent LLMs of different sizes. 
These models were chosen to assess the impact of model size on its robustness to the order of options in MCQ task.

\begin{table*}
\small
\centering
\begin{tabular}{lrrrrrr}
\toprule 
\multirow{2}{*}{\bf Tasks} & \multicolumn{3}{c}{\bf GPT-4}&  \multicolumn{3}{c}{\bf InstructGPT}\\
\cmidrule(lr){2-4}
\cmidrule(lr){5-7}
&Vanila&Min&Max&Vanila&Min&Max\\
\midrule
CSQA &84.3&\color{cadmiumred}{-12.6}&\color{cadmiumgreen}{+10.3}&72.3&\color{cadmiumred}{-24.0}&\color{cadmiumgreen}{+19.1}\\
Logical Deduction&92.3&\color{cadmiumred}{-8.1}&\color{cadmiumgreen}{+5.0}&64.0&\color{cadmiumred}{-39.4}&\color{cadmiumgreen}{+34.7}\\
Abstract Algebra &57.0&\color{cadmiumred}{-30.0}&\color{cadmiumgreen}{+23.0}&33.0&\color{cadmiumred}{-31.0}&\color{cadmiumgreen}{+39.0}\\
High School Chemistry &71.9&\color{cadmiumred}{-23.6}&\color{cadmiumgreen}{+18.2}&44.8&\color{cadmiumred}{-28.5}&\color{cadmiumgreen}{+38.0}\\
Professional Law &66.1&\color{cadmiumred}{-12.7}&\color{cadmiumgreen}{+12.1}&48.6&\color{cadmiumred}{-24.9}&\color{cadmiumgreen}{+25.7}\\
\bottomrule
\end{tabular}
\caption{\textbf{Zero-shot order sensitivity}; both GPT-4 and InstructGPT exhibit a notable level of sensitivity to the order of options across various benchmarks.}
\label{tab:attack-zero}
\end{table*}

\paragraph{data:} 
To investigate the sensitivity of LLMs to the order of options and the reasons behind this phenomenon, we conducted experiments on five distinct MCQ benchmarks. These benchmarks are as follows:
CSQA \citep{talmor2018commonsenseqa}: A commonsense multiple-choice question answering dataset, where each question is accompanied by 5 options. 
Abstract Algebra, High School Chemistry, and Professional Law from the MMLU benchmark \citep{hendrycks2020measuring}: These benchmarks consist of multiple-choice questions with 4 options provided for each question. 
And, Logical deduction from the Big-Bench dataset \citep{srivastava2022beyond}: This benchmark offers multiple-choice questions with 3 options for each question.
Our selection of these benchmarks was guided by three specific criteria:
(1) Domain diversity: We aimed to investigate the sensitivity to options order across different domains. 
(2) Varying option numbers: In order to explore the impact of the number of provided options, we selected benchmarks with different option counts, namely 3, 4, and 5 options per question. 
And (3) performance levels: By incorporating benchmarks with varying levels of LLMs' demonstrated performance, we sought to better understand how model proficiency influences sensitivity to the options order. 

\section{Sensitivity to Order}
In this section, we first investigation the sensitivity of InstructGPT and GPT-4 to the order of options in the zero-shot setting. 
Then, we set out to determine whether introducing demonstrations to the prompt in the few-shot setting can enhance the models' robustness.
To quantify sensitivity, we calculate the sensitivity gap, which is the difference between the maximum and minimum LLMs' performance when using an oracle ordering. In other words, we examine how specific reordering of options affects the models' predictions when the ground truth is known. 

\subsection{Zero-shot Sensitivity}
\label{sec:zer-sens}
The result of LLMs sensitivity to the order of options is presented in Table \ref{tab:attack-zero}. 
Several noteworthy observations emerge from these results: 
(1) GPT-4 demonstrates significantly lower sensitivity gap compared to InstructGPT. This suggests that GPT-4 is less affected by the rearrangement of options in the prompt, making it more robust in handling such variations.
(2) Even in tasks where GPT-4 achieves high accuracy levels exceeding $90\%$, we still observe a considerable sensitivity gap of $13.1\%$. This indicates that even high-performing models are susceptible to changes in options order, which can impact their fair assessment. 
(3) Although the sensitivity gap shows some correlation with the models' performance, tasks where LLMs perform poorly do not necessarily exhibit higher sensitivity gaps. This suggests that factors beyond overall accuracy may also influence LLMs' sensitivity to options order.
(4) The domain and the number of options in the MCQ tasks seem to affect the model's performance. However, we do not observe a clear correlation between these factors and the sensitivity gap. 

\begin{figure*}[th!]
    \centering
    \includegraphics[width=0.7\paperwidth]{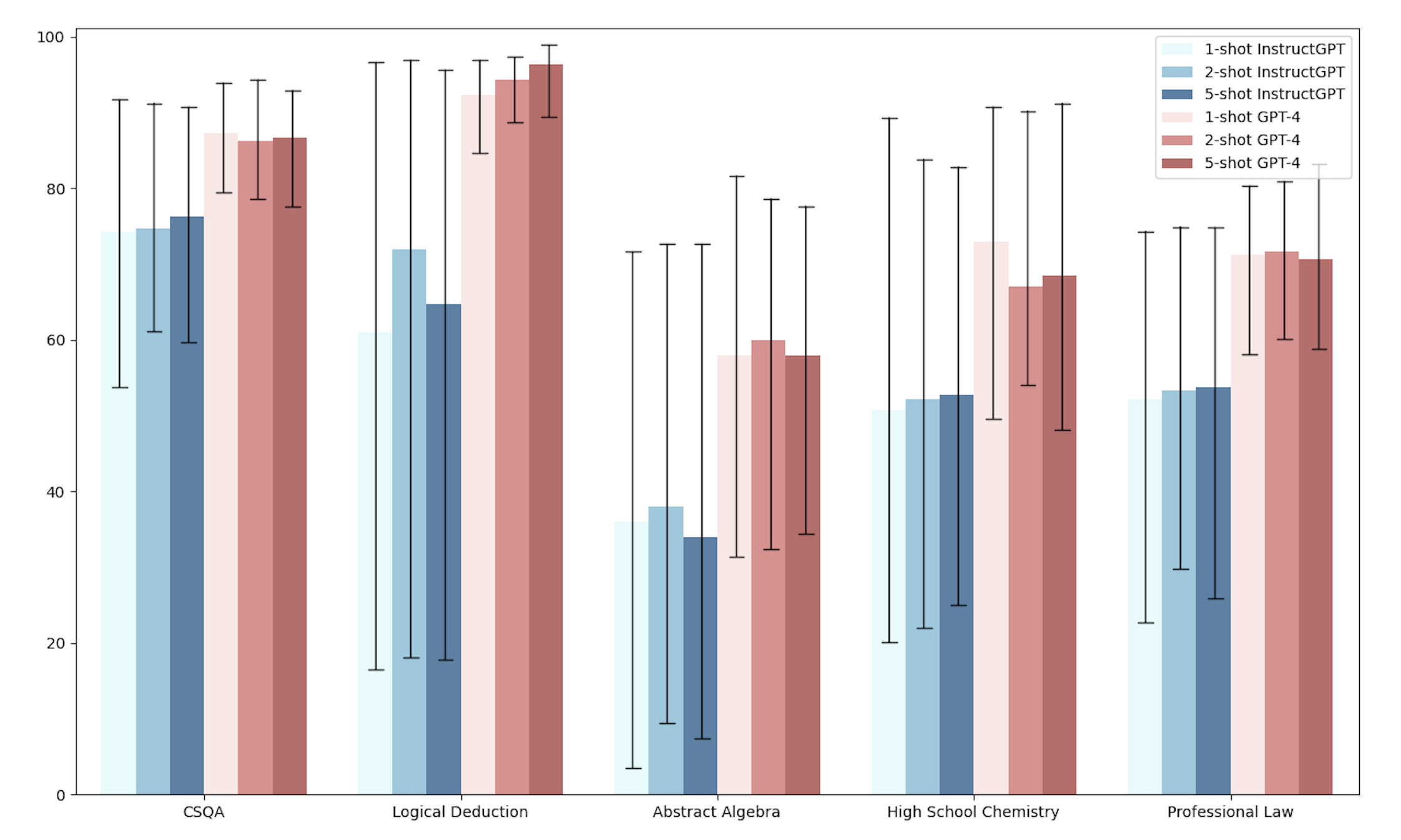}
    \caption{\textbf{Order sensitivity in few-shot setting:} The error bars represent the range of minimum and maximum accuracy achievable in each task through oracle reordering. Our observations are as follows: (1) The sensitivity gap consistently remains substantial even with the addition of more demonstrations in the few-shot setting. (2) As performances improve, the sensitivity gap shrinks. (3) Adding more demonstrations does not necessarily result in a reduction of the sensitivity gap.}
    \label{fig:few-sens}
\end{figure*}

\subsection{Can Demonstration in Few-shot Setting Resolve the Sensitivity?}
Having demonstrated the high level of sensitivity when zero-shot prompting LLMs, a crucial question that arises is whether adding demonstrations in the few-shot setting to the prompt can enhance the models' robustness. To address this, we select demonstrations in the few-shot setting by sampling the most similar instances. We achieve this by computing the Euclidean distance over vector representations of questions obtained from Sentence-RoBERTa \citep{reimers2019sentence}. 
The result of order sensitivity in the few-shot setting are visualized in Figure \ref{fig:few-sens} (more detailed results is provided in Appendix). 
Each bar in the figure is accompanied by error bars, representing the range of maximum and minimum model performance achievable by reordering the options, with knowledge of the ground truth. 
From the results, we make the following observations: Firstly, the sensitivity gap consistently remains substantial even with the inclusion of more demonstrations in the few-shot setting. 
Furthermore, as performances improve, the sensitivity gap tends to shrink. 
However, adding more demonstrations does not necessarily lead to a reduction in the sensitivity gap. 
This highlights the complexity of the relationship between model performance, the presence of demonstrations, and sensitivity to options order in the few-shot setting, i.e., while demonstrations may marginally improve robustness, they do not entirely mitigate the models' sensitivity to options order. 

\section{Why Do LLMs Show Sensitivity to the Order of Options?}
After analyzing instances in which reordering the options resulted in a change in LLMs prediction, we arrive at the following conjecture:
\begin{conjecture}
The sensitivity of LLMs to the order of options in multiple-choice questions arises from the interaction of two colluding forces:
(1) Uncertainty of LLMs regarding the correct answer among the top possible choices.
And (2) positional bias, leading LLMs to favor specific options based on the order they appear in, depending on the question.
\end{conjecture}
In the subsequent sections, we begin by empirically validating the conjecture. 
Furthermore, we identify specific patterns in the options that either amplify or mitigate the model's bias towards their placement.


\subsection{Uncertainty Meets Positional Bias}
To empirically validade our conjecture we devise qualitative experiments aimed at verifying each underlying reason behind the order sensitivity.

\paragraph{Uncertainty:} 
We assess the uncertainty of LLMs concerning instances where reordering affects predictions through a three-step analytical approach. It is important to highlight that GPT-4 and InstructGPT lack direct confidence measurements, necessitating our indirect analyses to validate our hypothesis.

\begin{figure}[th!]
    \centering
    \includegraphics[width=0.9\columnwidth]{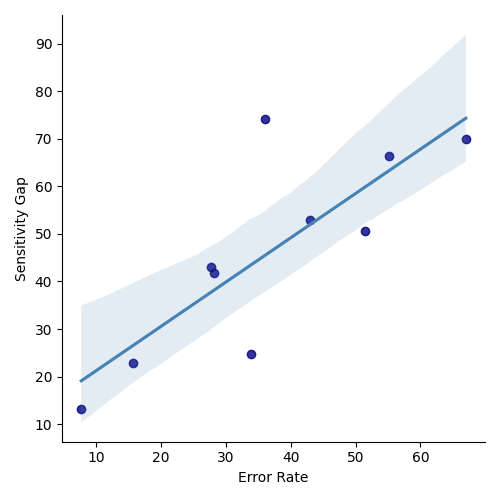}
    \caption{Correlation between the sensitivity gap and error rate for GPT-4 and InstructGPT across various multiple-choice question (MCQ) tasks.}
    \label{fig:cor}
\end{figure}

(1) The sensitivity gap, which comprises instances where reordering changes the prediction, exhibits a strong correlation with the error rate. The correlation plot between sensitivity gap and LLMs error rate on different benchmarks is depicted in Figure \ref{fig:cor}.

\begin{table*}[th!]
\small
\centering
\begin{tabular}{llrrrrrr}
\toprule 
&\multirow{2}{*}{\bf Tasks}& \multicolumn{3}{c}{\bf Sorted Options}&  \multicolumn{2}{c}{\bf \# Options}\\
\cmidrule(lr){3-5}
\cmidrule(lr){6-8}
&&Hits@1&Hits@2&Hits@3&Top-2&Top-3&All\\
\midrule
\multirow{5}{*}{\rotatebox[origin=c]{90}{\bf GPT-4}}&CSQA &81.3&95.1&98.2&84.2&85.1&84.3\\
&Logical Deduction&85.3&95.7&97.9&94.8&92.3&92.3\\
&Abstract Algebra &55.0&72.0&88.0&57.0&52.0&57.0\\
&High School Chemistry &64.0&74.4&76.8&65.5&68.1&71.9\\
&Professional Law &51.7&62.9&74.1&65.3&65.1&66.1\\
\midrule
\multirow{5}{*}{\rotatebox[origin=c]{90}{\bf InstructGPT}}&CSQA &63.4&82.3&90.3&70.6&72.1&72.3\\
&Logical Deduction&65.6&93.0&97.6&66.2&64.0&64.0\\
&Abstract Algebra &28.0&52.0&73.0&26.0&29.0&33.0\\
&High School Chemistry &30.0&51.7&66.9&37.9&40.1&44.8\\
&Professional Law &40.0&63.3&76.7&47.7&50.6&48.6\\
\bottomrule
\end{tabular}
\caption{Assessing sorting options quality with LLMs and analyzing the impact of reducing options complexity on models performance.}
\label{tab:hits}
\end{table*}

(2) More than $60\%$ of the sensitive samples identified in GPT-4 also exhibit sensitivity in InstructGPT. 
Additionally, many of these instances contain semantically similar top-choice options. 
For instance, both InstructGPT and GPT-4 consider the question ``{\promptfont Most items in retail stores are what even when they are on sale? A) overpriced B) purchase C) expensive D) park E) buying}'' from CSQA with top choices ``overpriced'' and ``expensive'', as sensitive.

(3) To further verify models' uncertainty towards sensitive instances, we conduct a self-verification process by posing the following question to the LLMs: ``{\promptfont Can more than one of the choices be a highly probable answer to the question? Please respond with `yes' or `no'. Question: \{question\}. Choices: \{options\}. Answer:}''. Remarkably, LLMs consistently predict "yes" for over $94\%$ of the sensitive cases across various benchmarks, further confirming their uncertainty in these scenarios.


\paragraph{Positional Bias:}  
We aim to explore the effect of positional bias in LLMs' order sensitivity by reducing sample difficulty, retaining only the top possible choices while preserving their original order of appearance, and eliminating the rest of the options. 
The goal is to isolate the influence of positional bias, disentangling it from other potential hidden factors impacting order sensitivity. 
To identify the top possible choices for each question, we ask LLMs to sort the options in descending order of probability for answering the question. 
We observe that the Hits@1 metric, which measures the accuracy of the gold truth being the first item in the sorted options, closely aligns with LLMs' overall task accuracy. 
Moreover, over $95\%$ and $100\%$ of instances that LLMs predict correctly are captured in Hits@2 and Hits@3, respectively. 
The detailed results of Hits@ metrics for both GPT-4 and InstructGPT are provided in Table \ref{tab:hits}.

With the successful identification of the top possible choices by asking LLMs to sort the options, we proceed to investigate the impact of removing the least probable choices on the models' performance, aiming to establish the presence of positional bias. 
The results of retaining only the top-2 and top-3 choices after sorting the options using LLMs themselves, while preserving their original order of appearance, are presented in Table \ref{tab:hits}. 
We observe that despite achieving high Hits@2 and Hits@3 scores (covering all the samples where models initially predicted them correctly), LLMs' performance remains nearly unchanged or exhibits incremental improvements or declines. 
This observation provides further evidence of the impact of positional bias in order sensitivity.

\begin{table}[t]
\small
\centering
\begin{tabular}{llrrrr}
\toprule 
&\multirow{2}{*}{\bf Tasks}& \multicolumn{2}{c}{\bf Amplify}&  \multicolumn{2}{c}{\bf Mitigate}\\
\cmidrule(lr){3-4}
\cmidrule(lr){5-6}
&&Pattern&Ord&Pattern&Ord\\
\midrule
\multirow{3}{*}{\rotatebox[origin=c]{90}{\bf GPT-4}}&5-option&2&AE&3&BA\\
&4-option&2&BD&1&AB\\
&3-option&2&AC&3&CB\\
\midrule
\multirow{3}{*}{\rotatebox[origin=c]{90}{\bf Inst}}&5-option&4&EA&1&BC\\
&4-option&4&EA&3&CB\\
&3-option&4&CA&3&CB\\
\bottomrule
\end{tabular}
\caption{\textbf{Optimal patterns and their best order instantiation} for amplifying and mitigating positional bias in different LLMs based on available number of options in multiple-choice questions.}
\label{tab:patt}
\end{table}
\begin{table*}[th!]
\small
\centering
\begin{tabular}{lrrrr}
\toprule 
\multirow{2}{*}{\bf Tasks}& \multicolumn{2}{c}{\bf GPT-4}&  \multicolumn{2}{c}{\bf InstructGPT}\\
\cmidrule(lr){2-3}
\cmidrule(lr){4-5}
&Amplifying-Bias& Mitigating-Bias&Amplifying-Bias& Mitigating-Bias\\
\midrule
CSQA &62.9&22.7&71.7&38.3\\
Logical Deduction&42.0& 10.1& 61.7& 0.9\\
Abstract Algebra &52.8&15.1&35.7&25.7\\
High School Chemistry &21.5&22.9&25.7&25.7\\
Professional Law &31.5&9.7&20.1&25.9\\
\bottomrule
\end{tabular}
\caption{\textbf{Percentage of initial sensitivity gap covered} using the identified patterns to amplify and mitigate positional bias. A higher percentage in amplifying bias and a lower percentage in mitigating bias indicate better performance in this context.}
\label{tab:favor}
\end{table*}
\subsection{What Patterns Amplify or Mitigate the Positional Bias?}
In here we investigate the impact of certain patterns in the options on the intensity of positional bias. We categorize our findings based on number of options and the target large language model. 
We limit our investigation to the order of the top-2 choices (extracted from the sorted options list) in the options and their impact on the models' prediction to identify influential patterns. 
We defer further analysis of patterns involving options beyond the top-2 choices to future research.

Our goal is to identify patterns that amplify the positional bias, increasing the probability of the LLM to choose one answer over another based on their position, or mitigating the positional bias, decreasing dependency of the LLM to choose one answer over another based on their position.  
Upon investigating the order and placement of top-2 choices in instances where reordering change the prediction, we discover four different patterns: 

\begin{itemize}
    \item \textbf{Pattern 1:} First choice in top-2 appear \textit{earlier} than the second choice in the options, and having \textit{less} gap (less number of other choices) between them helps the goal more, i.e., to amplify or mitigate the positional bias. 
    \item \textbf{Pattern 2:} First choice in top-2 appear \textit{earlier} than the second choice in the options, and having \textit{more} gap between them helps the goal more.
    \item \textbf{Pattern 3:} First choice in top-2 appear \textit{later} than the second choice in the options, and having \textit{less} gap between them helps the goal more.
    \item \textbf{Pattern 4:} First choice in top-2 appear \textit{later} than the second choice in the options, and having \textit{more} gap between them helps the goal more.
\end{itemize}


The best pattern, along with its best corresponding order instantiation (placement of top-2 choices), for amplifying or mitigating positional bias based on the type of LLMs and the number of options in the multiple-choice question task is presented in Table \ref{tab:patt}. 
For instance, to amplify the positional bias between two choices with the objective of increasing the probability of selecting the first choice as the answer for GPT-4, pattern number 2 proves to be the most effective. 
The ideal instantiation of this pattern is to place the first choice in option A and the second choice in option E.
Investigating the positional bias in various LLMs with different numbers of options in the MCQ task reveal interesting findings. In both GPT-4 and InstructGPT, the most influential pattern to amplify the bias remains the same while for mitigating bias the best pattern jumps between first and third patterns. 
Furthermore, there is a notable contrast between InstructGPT and GPT-4 in their reactions to patterns regarding the order of appearance in the top-2 choices or the size of the gap between them.  
In overall, to mitigate bias, it appears to be more effective for the top-2 choices to either appear in the first two options or in the second and third options. Conversely, for amplifying bias, it is preferable for the top-2 choices to be positioned in the first and last options. 

To validate the discovered patterns and their impact on LLMs' order sensitivity, we conducted two sets of experiments. 
Firstly, to confirm the effectiveness of patterns amplifying positional bias, we selected the best instantiation of each pattern and measured the performance improvement achieved by placing only the top-2 choice (where the ground truth is at top-1, and top-2 is obtained by sorting the options) in that instantiation. Meanwhile, we kept the order of appearance for other choices. 
Also, we measured the decrease in LLMs' performance by using the reverse instantiation of the pattern. 
Our goal here, is to assess the extent to which the sensitivity gap identified in Section \ref{sec:zer-sens} could be achieved simply by utilizing the most impactful placement. As a result, a higher percentage of coverage over the original sensitivity gap here means that the identified pattern did a better job at amplifying bias.

Secondly, to validate the patterns mitigating the bias, we performed a similar experiment as in Section \ref{sec:zer-sens}, but this time, we fixed the top-2 choices in the placements provided in Table \ref{tab:patt} and reordered all other options accordingly. 
The purpose here is to demonstrate how much of the sensitivity gap can be minimized by following identified mitigating patterns. As a result, a lower percentage of coverage over the original sensitivity gap here means that the identified pattern did a better job at mitigating bias.

Table \ref{tab:favor} presents the percentage of initial sensitivity gap covered (initial sensitivity gaps are from Table \ref{tab:attack-zero}) by the optimal pattern for amplifying and mitigating positional bias, with more detailed results available in Appendix. 
A higher percentage in amplifying bias and a lower percentage in mitigating bias indicate better performance of the identified pattern. 
The amplifying patterns demonstrate sensitivity gap coverage ranging from $20\%$ to $72\%$, while the mitigating bias pattern ranges from $0.9\%$ to $38\%$. 
These results validate the effectiveness of the identified pattern for both amplifying and mitigating bias. 
Additionally, in most cases, the amplifying pattern covers a considerably greater portion of the sensitivity gap comparing to the mitigating pattern. 
It is important to highlight that the patterns we have identified for amplifying bias can serve as valuable insights for enhancing model performance or launching adversarial attacks against them. Furthermore, the patterns we have established for mitigating bias can play a crucial role in shaping benchmark design and guiding annotating efforts to create less biased benchmarks.

\begin{table*}
\small
\centering
\begin{tabular}{lrrrr}
\toprule 
\bf Tasks& \multicolumn{2}{c}{\bf GPT-4}&  \multicolumn{2}{c}{\bf InstructGPT}\\
\cmidrule(lr){2-3}
\cmidrule(lr){4-5}
&Majority&MEC&Majority&MEC\\
\midrule
CSQA &86.1 \color{cadmiumgreen}{($+1.8$)}& 81.2 \color{cadmiumred}{($-3.1$)}&74.7 \color{cadmiumgreen}{($+2.4$)}&67.3 \color{cadmiumred}{($-5.0$)}\\
Logical Deduction&94.3 \color{cadmiumgreen}{($+2.0$)}&97.4 \color{cadmiumgreen}{($+5.1$)}&72.0 \color{cadmiumgreen}{($+8.0$)}& 57.1 \color{cadmiumred}{($-6.9$)}\\
Abstract Algebra &57.0 ($0.0$)& 59.0 \color{cadmiumgreen}{($+2.0$)}&38.0 \color{cadmiumgreen}{($+5.0$)} & 31.0 \color{cadmiumred}{($-2.0$)}\\
High School Chemistry &71.9 ($0.0$)&77.2 \color{cadmiumgreen}{($+5.3$)}&45.8 \color{cadmiumgreen}{($+1.0$)} &39.4 \color{cadmiumred}{($-5.4$)}\\
Professional Law &67.3 \color{cadmiumgreen}{($+1.2$)}&66.3 \color{cadmiumgreen}{($+0.2$)}&54.3 \color{cadmiumgreen}{($+5.7$)}&47.2 \color{cadmiumred}{($-1.4$)}\\
\bottomrule
\end{tabular}
\caption{Impact of calibration methods on LLMs' performance.}
\label{tab:calib}
\end{table*}

\section{Calibrating LLMs for MCQ Tasks}
We conduct an in-depth investigation into how large language models react to changes in the order of options, and investigate the reasons behind their sensitivity to such changes. 
Through our exploration, we have clearly observed that LLMs are highly responsive to the sequence in which options are presented. 
This has led us to a critical juncture where we need to focus on methods to improve the models' resilience to variations in options order, ensuring more dependable and trustworthy evaluations.

One potential solution we have considered is the calibration of LLMs predictions. The outcomes of calibrating LLMs predictions to mitigate order sensitivity by taking majority vote over models prediction in 10 random reorders in a simple bootstrapping approach \citep{stickland2020diverse, hou2023large}, are provided in Table \ref{tab:calib}. 
Our analysis has unveiled a significant observation: employing a majority vote approach for evaluating LLMs across a range of benchmarks results in a substantial performance improvement of up to 8 percentage points. 
Notably, we find that GPT-4 showcases a higher level of stability against changes in options order compared to InstructGPT, a trait that persists even after implementing prediction calibration. 
Furthermore, while LLMs' performance on benchmarks featuring four options might be somewhat inferior to those with three or five options, GPT-4 displays a greater resilience following prediction calibration. 
In contrast, InstructGPT demonstrates heightened robustness in specific contexts like CSQA and high school chemistry, where the effects of calibration are more pronounced. 

We have also incorporated the approach of Multiple Evidence Calibration (MEC) introduced by \citet{wang2023large}. 
In their work, they propose to counteract LLMs' sensitivity by prompting the model to generate an explanation before providing its prediction. 
We adopt their provided prompt for solving MCQ tasks. 
The impact of applying MEC calibration and its implications for order sensitivity are outlined in Table \ref{tab:calib}. 

The results from InstructGPT performance reveal that the introduction of MEC calibration results in a consistent decrease in model performance. This behavior contradicts the outcomes achieved through majority voting and underscores the unsuitability of MEC calibration for multiple-choice question tasks.
In the case of GPT-4, the integration of MEC calibration also yields contrasting outcomes with respect to majority voting, particularly evident in benchmarks such as CSQA, abstract algebra, and high school chemistry. For logical deduction and professional law benchmarks, while both majority voting and MEC calibration result in improving the model performance, the amount of improvement differs considerably, thus casting doubt on the reliability of the MEC approach in GPT-4 as well. 
Subsequent analysis of LLMs' mispredictions following MEC calibration unveils a noteworthy trend: the act of asking the models to explain their reasoning amplifies their uncertainty (partly due to hallucination), particularly in instances where the model's initial confidence was not very high. 
This, in turn, either frequently guides the models to opt for answer among the top choices, which is positioned earlier in the available options or causing the model to not picking any final answer.

\section{Related Work}
Large language models (LLMs) show remarkable accomplishments and capabilities on various NLP tasks, including answering multiple-choice questions. 
In order to ascertain the dependability of LLMs' proficiency, it becomes imperative to delve into the robustness of their performance when subjected to subtle changes in the input.

\paragraph{LLMs and multiple-choice questions}  
In recent years, multiple-choice questions have been introduced as a method for assessing the reasoning and fact-retrieval capabilities of models \citep{richardson2013mctest,talmor2018commonsenseqa,clark2020f,hendrycks2020measuring}. Despite the intricate nature of these tasks, significant strides have been made by large language models achieving human-like performances across various MCQ benchmarks \cite{lievin2022can,robinson2022leveraging,openai2023gpt-4,savelka2023large,anil2023palm}. 
However, the ability of these tasks to effectively gauge the reasoning and factual knowledge of LLMs, along with the reliability of the evaluation settings, presents substantial challenges that warrant deeper investigation.

\paragraph{Sensitivity of LLMs} 
With the growing prominence of LLMs in addressing NLP tasks, significant attention has been devoted to examining the robustness and vulnerabilities of these models. These efforts predominantly focus on two distinct levels:
(1) At the instance level, researchers investigate the robustness of LLMs by studying how modifications or adversarial attacks impact individual instances. For example, \citet{zhao2021calibrate} reveal LLMs' sensitivity to prompt choice and demonstrations order in in-context learning (ICL). 
\citet{hou2023large} show LLMs are sensitive to the order of sequential interaction histories when used as conditions in ranking candidates for recommender systems. 
\citet{wang2023adversarial} launch adversarial attacks on LLM predictions through modifications to ICL demonstrations. \citet{wang2023large} also explore LLMs' susceptibility to the order of response appearances from candidate models when LLMs serve as referees. 
(2) At the alignment level, attempts are made to deliberately misalign LLMs to manipulate their behavior, often referred to as "jailbreaking." \citet{perez2022ignore,zou2023universal} achieve misalignment by adversarially attacking the prompt. In a similar vein, \citep{wolf2023fundamental} propose a theoretical framework that exposes limitations in aligning LLMs, demonstrating there exist prompts that can cause models to exhibit any behavior with finite probability. Furthermore, \citet{wei2023jailbroken} propose that jailbreaking arises from conflicting objectives and mismatched generalization, utilizing their hypothesis to develop effective jailbreak strategies.

\section{Discussion and Conclusion}
We investigate the inherent sensitivity of language models to the arrangement of options in multiple-choice questions. Upon measuring the intensity of LLMs sensitivity, our aim was twofold: to pinpoint the underlying source of this sensitivity and propose potential solutions to enhance the models' robustness. 
Our evaluations unequivocally reveal that LLMs not only exhibit pronounced sensitivity to options order, but also that this sensitivity diminishes only slightly when demonstrations are integrated into the few-shot setting under specific circumstances.

While our primary focus has been on multiple-choice questions, we have also detected a parallel phenomenon—albeit with varying degrees of sensitivity—in other tasks involving multiple fragments (e.g. the options in MCQ) within inputs. 
This encompasses tasks like odd word detection, sorting lists of items, and ranking documents. While these observations have been noted, further exploration into these tasks is reserved for future efforts.

In seeking to uncover the root cause of order sensitivity, we conjecture that the issue arises from LLMs' positional bias, particularly manifesting in uncertain instances. 
We verify our conjecture by conducting diverse experiments that highlight impactful patterns that either magnify or mitigate this positional bias. 
Despite the validation provided through detailed experimentation, a deeper comprehension of the issue's origin necessitates a thorough exploration of the training data.

To improve the robustness of LLMs' sensitivity against options order, we consider two calibration techniques: majority vote and multiple evidence calibration (MEC). 
While both methods display promising outcomes, contributing to the improvement of model performance, they are not without their respective limitations. 
Majority voting is computationally expensive, while MEC diverges significantly from majority voting, casting doubts on its applicability to MCQ tasks.
As a result, in order to establish a reliable and accurate evaluation framework for LLMs in the context of multiple-choice questions, it is imperative to develop more efficient calibration strategies. Moreover, refining the evaluation metrics holds the potential to improve LLMs' ability to withstand the challenges posed by options order sensitivity. These avenues present opportunities for in-depth exploration in future works.

\section*{Acknowledgements}
We would like to thank Sameer Singh for his valuable comments. 



\bibliography{main}
\bibliographystyle{acl_natbib}

\appendix

\section{Detailed Results}
\label{sec:appendix}
The detailed result of order sensitivity in few-shot setting is provided in Tables \ref{tab:instruct-few} and \ref{tab:gpt4-few} for InstructGPT and GPT-4 respectively. 
Moreover, we present the impact of the identified patterns aimed at amplifying and mitigating positional bias on order sensitivity in Table \ref{tab:patt-det}.

\begin{table*}[t!]
\small
\centering
\begin{tabular}{lrrrrrrrrr}
\toprule 
\multirow{2}{*}{\bf Tasks} & \multicolumn{3}{c}{\bf 1-shot}&  \multicolumn{3}{c}{\bf 2-shot}&  \multicolumn{3}{c}{\bf 5-shot}\\
\cmidrule(lr){2-4}
\cmidrule(lr){5-7}
\cmidrule(lr){8-10}
&Vanila&Min&Max&Vanila&Min&Max&Vanila&Min&Max\\
\midrule
CSQA &74.2&53.4&92.1&74.7&60.7 & 91.6&76.3&59.3&91.1\\
Logical Deduction&61.0&16.0&97.0&72.0&17.7&97.3&64.7&17.3&96.0\\
Abstract Algebra &36.0&3.0&72.0&38.0&9.0&73.0&34.0&7.0&73.0\\
High School Chemistry &50.7&19.7&89.7&52.2&21.6&84.2&52.7&24.6&83.2\\
Professional Law &52.1&22.3&74.7&53.3&29.3&75.3&53.7&25.5&75.2\\
\bottomrule
\end{tabular}
\caption{Few-shot order sensitivity in InstructGPT.}
\label{tab:instruct-few}
\end{table*}
\begin{table*}[t!]
\small
\centering
\begin{tabular}{lrrrrrrrrr}
\toprule 
\multirow{2}{*}{\bf Tasks} & \multicolumn{3}{c}{\bf 1-shot}&  \multicolumn{3}{c}{\bf 2-shot}&  \multicolumn{3}{c}{\bf 5-shot}\\
\cmidrule(lr){2-4}
\cmidrule(lr){5-7}
\cmidrule(lr){8-10}
&Vanila&Min&Max&Vanila&Min&Max&Vanila&Min&Max\\
\midrule
CSQA &87.2&79.1&94.3&86.3&78.2&94.7&86.7&77.2&93.3\\
Logical Deduction&92.3&84.3&97.3&94.3&88.3&97.7&96.3&89.0&99.3\\
Abstract Algebra &58.0&31.0&82.0&60.0&32.0&79.0&58.0&34.0&78.0\\
High School Chemistry &72.9&49.2&91.1&67.1&53.7&90.6&68.5&47.7&91.6\\
Professional Law &71.2&57.7&80.7&71.7&59.7&81.3&70.6&58.4&83.6\\
\bottomrule
\end{tabular}
\caption{Few-shot order sensitivity in GPT-4.}
\label{tab:gpt4-few}
\end{table*}

\begin{table*}[t!]
\small
\centering
\begin{tabular}{lrrrrrrrr}
\toprule 
\multirow{2}{*}{\bf Tasks}& \multicolumn{4}{c}{\bf GPT-4}&  \multicolumn{4}{c}{\bf InstructGPT}\\
\cmidrule(lr){2-5}
\cmidrule(lr){6-9}
&\multicolumn{2}{c}{\bf Amplifying-Bias}&  \multicolumn{2}{c}{\bf Mitigating-Bias}&\multicolumn{2}{c}{\bf Amplifying-Bias}&  \multicolumn{2}{c}{\bf Mitigating-Bias}\\
\cmidrule(lr){2-3}
\cmidrule(lr){4-5}
\cmidrule(lr){6-7}
\cmidrule(lr){8-9}
&Min&Max&Min&Max&Min&Max&Min&Max\\
\midrule
CSQA &-8.0& +6.4& -4.8 & +0.4&-16.0&+14.9&-7.7&+8.8\\
Logical Deduction&-3.1&+2.4& +1.3 & +1.3&-28.4&+17.3& +0.7&+0.7\\
Abstract Algebra &-19.0&+9.0&-7.0&+1.0&-17.0&+8.0&-9.0&+9.0\\
High School Chemistry &-7.0& +2.0& -11.6&-2.0&-11.6 & +5.5&-9.3&+7.8\\
Professional Law &-3.8& +4.0& +3.2&+5.6&-6.4& +3.7 &-7.6&+5.5\\
\bottomrule
\end{tabular}
\caption{Sensitivity gap after applying the identified patterns to amplify and mitigate positional bias.}
\label{tab:patt-det}
\end{table*}

\end{document}